\documentclass{article}

\PassOptionsToPackage{numbers, compress}{natbib}
\usepackage[preprint]{neurips_2020}

\usepackage[utf8]{inputenc} 
\usepackage[T1]{fontenc}    
\usepackage{hyperref}       
\usepackage{url}            
\usepackage{booktabs}       
\usepackage{amsfonts}       
\usepackage{nicefrac}       
\usepackage{microtype}      
\usepackage{amsmath}

\usepackage{gensymb}
\usepackage{multirow}
\usepackage{booktabs}
\usepackage{pifont}
\usepackage{arydshln}
\usepackage{graphicx}
\usepackage{caption}
\usepackage{subcaption}
\usepackage{xspace}
\usepackage{xcolor}
\usepackage{dashbox}
\usepackage{enumitem}
\usepackage{algorithm}
\usepackage{algpseudocode}

\algrenewcommand\algorithmicrequire{\textbf{Input:}}
\algrenewcommand\algorithmicensure{\textbf{Step}}

\makeatletter
\newcommand{\printfnsymbol}[1]{%
  \textsuperscript{\@fnsymbol{#1}}%
}
\makeatother

\title{Few-Shot Learning with Intra-Class Knowledge Transfer\\
}

\author{%
  Vivek Roy\thanks{Equally contributed.} \\
  Carnegie Mellon University\\
  \texttt{vroy@cs.cmu.edu} \\
   \And
   Yan Xu\printfnsymbol{1} \\
  Carnegie Mellon University\\
  \texttt{yxu2@cs.cmu.edu} \\
   \And
   Yu-Xiong Wang \\
  Carnegie Mellon University\\
  \texttt{yuxiongw@cs.cmu.edu} \\
   \And
   Kris Kitani \\
  Carnegie Mellon University\\
  \texttt{kkitani@cs.cmu.edu} \\
   \And
   Ruslan Salakhutdinov \\
  Carnegie Mellon University\\
  \texttt{rsalakhu@cs.cmu.edu} \\
   \And
   Martial Hebert \\
  Carnegie Mellon University\\
  \texttt{hebert@cs.cmu.edu} \\
}

\begin{document}

\maketitle

\begin{abstract}

We consider the few-shot classification task with an unbalanced dataset, in which some classes have sufficient training samples while other classes only have limited training samples.
Recent works have proposed to solve this task by augmenting the training data of the few-shot classes using generative models with the few-shot training samples as the seeds.
However, due to the limited number of the few-shot seeds, the generated samples usually have small diversity, making it difficult to train a discriminative classifier for the few-shot classes.
To enrich the diversity of the generated samples, we propose to leverage the intra-class knowledge from the neighbor many-shot classes with the intuition that neighbor classes share similar statistical information.
Such intra-class information is obtained with a two-step mechanism.
First, a regressor trained only on the many-shot classes is used to evaluate the few-shot class means from only a few samples.
Second, superclasses are clustered, and the statistical mean and feature variance of each superclass are used as transferable knowledge inherited by the children few-shot classes.
Such knowledge is then used by a generator to augment the sparse training data to help the downstream classification tasks.
Extensive experiments show that our method achieves state-of-the-art across different datasets and $n$-shot settings.

\end{abstract}
\section{Introduction}


Machine learning algorithms have recently achieved remarkable performance in visual recognition tasks, benefiting from the knowledge that learned from a massive amount of training examples.  However, when the training data is limited, performance will drop dramatically.  As a comparison, human learners can understand the concept of an object given even one example and can still generalize reasonably well to novel instances~\cite{burt1933mind}.  Such an ability to learn from only a few training samples, or the ability of \textit{few-shot learning}, is crucial in the development of human-level artificial intelligence.  To take one step towards this direction, we present here a learning-based method to mimic the few-shot learning process of humans.  Our model transfers meaningful knowledge learned from many-shot classes to few-shot classes and then uses the knowledge to augment data for few-shot classes with a deep generative model.

\begin{figure*}[t]
\centering
 \includegraphics[width=\textwidth]{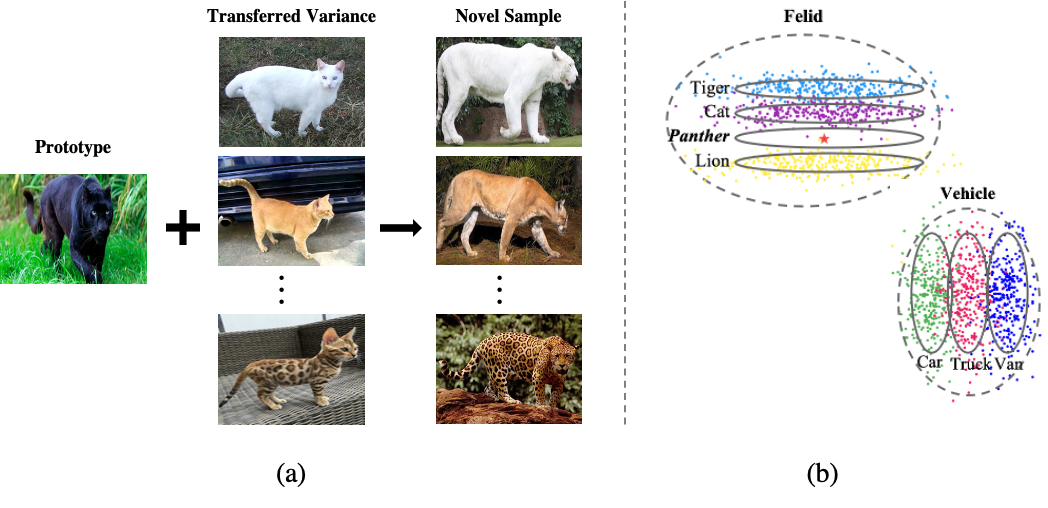}
 \caption{(a) Given one simple or a prototype of a few-shot class, humans can easily imagine what other novel samples would look like using the knowledge learned from other many-shot classes. (b) When the classes are close to each other in the feature dimension, their statistical mean and variance are similar.  When the classes are far away, their statistical mean and variance are different.\vspace{-15pt}}
 \label{fig:teaser}
\end{figure*}

What kind of knowledge would be both informative and generalizable?  To categorize an object, humans, at a minimum, require the information about the category's mean and variance along each dimension in an appreciate feature space~\cite{salakhutdinov2012one}.  The mean represents a prototype of what an object from this category should generally look like, and the variance sets an appreciate changing range for the feature of each dimension. With few-shot samples, humans can summarize a mean from the samples, but the variance seems difficult to estimate.  To obtain the variance, humans usually transfer previously learned knowledge from other related many-shot classes and use this knowledge as inductive information for estimating the feature variances of few-shot classes.
For instance, from a picture of a black panther (few-shot), humans can imagine what white or yellow panthers look like by generalizing the color variance learned from cats (many-shot), as illustrated in Figure \ref{fig:teaser}a.  Such a generalization is reasonable since both panther and cat belong to the same felid super-category, and they share similar variances in specific feature dimensions. \textit{e.g.} color, shape, \textit{etc}.  However, such knowledge transferring is only reasonable when the many-shot class and the few-shot class are close in feature dimensions.  When the many-shot class is far away from the few-shot class, such a knowledge transferring could be useless or even wrong.  For example, the shapes of vehicles should not be applied to estimate the shapes of animals, as shown in Figure \ref{fig:teaser}b.

Our method leverages such intra-class knowledge transferring to estimate the mean and feature variance of few-shot classes.  However, instead of transferring knowledge from the neighbor classes, we make the few-shot classes inherit knowledge from their superclasses with a two-step mechanism.  In the first step, for each few-shot classes, we use a neural network regressor trained on the many-shot classes to estimated from few-shot samples their class means.  For many-shot classes, we directly compute the class mean from the samples.  In the second step, we cluster the means of both many-shot and few-shot classes.  Based on the clustering result, we then compute for each superclass the mean and variance, which will inherit by the children few-shot classes as the transferred knowledge for solving downstream few-shot classification tasks.

The few-shot classification task is generally solved under the meta-learning~\cite{thrun1998lifelong} frame. To deal with the lack of training samples for few-shot classes, many recently proposed approaches augment the training data with a meta-generator using the few-shot samples as the generation seed~\cite{wang2018low, wang2019meta, xian2019f, zhang2019few}.  However, data augmentation using only the few-shot samples is not sufficient.  First, the few-shot samples could lie close to the class boundary in the feature dimension, which makes the generated samples extremely biased.  Second, it is difficult for the generator to have a reasonable estimation of the feature variance given only a few training samples, which makes the generated samples less diverse.  Instead of augmenting the training data using only the few-shot samples, our approach first estimates the class mean and variance with the two-step mechanism mentioned above.  It then uses transferred knowledge together with the few-shot samples to augment the few-shot data.  Extensive experiments show that our approach has achieved state-of-the-art (SOTA) performance across several few-shot learning benchmark datasets.

We aim to use the proposed approach to mimic the human few-shot learning process, in which the knowledge transferring helps obtain an estimation on the prototype and the feature variance, and the meta-generator helps imagine what novel samples look like.  We summarize our contribution as two folds.  First, we proposed a two-step mechanism estimate the class means and feature variances of the few-shot classes using only the given data, without the requirement of extra information injection.  Second, we leverage the information transferred from many-shot classes to augment the sparse training data for few-shot classes using a meta-generator.

\section{Related Work}
\label{sec:related}

Few-shot learning~\cite{MillerCVPR2000,FeiFeiTPAMI2006} is one of the most important yet unaddressed problems in machine learning.  Recently, to leverage the excellent expression ability of deep neural networks, many deep learning-based methods~\cite{dvornik2019diversity,allen2019infinite,li2019lgm,yoon2019tapnet,dhillon2020baseline,LakeScience2015,santoro2016one,HariharanICCV2017,triantafillou2017few,mishra2018asimple,douze2018low,wang2018low,chen2019acloser,triantafillou2020meta} have been proposed for solving few-shot learning problems.  These methods accumulate generalizable knowledge from previously seen tasks~\cite{baxter1997bayesian} and then apply the knowledge to novel tasks to speed up the learning procedure and achieve better performance.  Many of these methods are within the paradigm of meta-learning or, learning to learn~\cite{bengio1992optimization}.  Generally, they sample a series of few-shot learning tasks from given base classes, explicitly learn and accumulate task-agnostic meta-knowledge and apply to novel tasks.  Variety types of such meta-knowledge have been explored.  Some methods propose to learn a generic feature embedding, mapping from the input space to a metric space, in which classification can be easily conducted using distance-based classifiers~\cite{KochICMLW2015,VinyalsNIPS2016,SnellArxiv2017,sung2018learning,ren2018meta,oreshkin2018tadam}.  Other methods treat the initialization of network parameters or the gradient updating rules as such meta-knowledge~\cite{andrychowicz2016learning,RaviICLR2017,munkhdalai2017meta,li2017meta,rusu2019meta, FinnICML2017,nichol2018reptile,finn2018probabilistic}.  There are also methods that learn a transferable strategy to predict model parameters according to a few class samples~\cite{WangECCV2016,wang2017learning, BertinettoNIPS2016,qiao2018few,qi2018low,gidaris2018dynamic}.  As a complementary to these discriminative methods, we propose a generative method.   Our method augments the sparse data using synthetic samples under the meta-learning framework.

Research in data synthesis has achieve encouraging progress in recent years~\cite{Salakhutdinov2012,LakeScience2015, GoodfellowNIPS2014,DixitCVPR2017,HariharanICCV2017,wang2018low,gao2018low,schwartz2018delta,zhang2018metagan}.  They use deep generative models to estimate the data distribution and generate samples that follow the sample distribution.  Modern generative models can synthesize high-quality images in terms of realism~\cite{brock2018large, song2019generative, karras2017progressive}.  However, using data synthesis for solving few-shot learning tasks is still challenging.  First, an improvement in the realism of synthetic samples does not equal to the gain on the performance of recognition tasks.  Shmelkov \textit{et al.} ~\cite{shmelkov2018good} shows that images generated by Generative Adversarial Networks (GANs), despite their impressive visual quality, does not necessarily help in improving the performance for classification tasks.  Second, it is difficult to capture the distribution of the whole class from just a few samples~\cite{SalimansNIPS2016}.  As a consequence, downstream models trained on the synthetic samples will be biased.  In our work, instead of synthesizing images using realism as an evaluation metric, we directly guide the image generation process using the classification objective in an end-to-end fashion.

The work that most closely related to ours is the one from Wang \textit{et al.}~\cite{wang2018low}.  They proposed a data hallucination method based on meta-learning, which directly uses few-shot classification accuracy to guide the learning of a generator in an end-to-end manner.  This generator takes a random noise and a sample from the few-shot data as seeds to synthesize more samples.  Another work similar to Wang \textit{et al.}~\cite{wang2018low} is Delta-encoder~\cite{schwartz2018delta}.  It learns to extract transferable intra-class deformations, or "deltas", between
same-class pairs of training examples, and to apply those deltas to the few provided examples of a novel class to efficiently synthesize samples for that new class.  The training of Delta-encoder is also directly guided by a downstream classifier.  Another work that leverages a similar idea is MetaGAN~\cite{zhang2018metagan}, which trains the classifier in an adversarial manner using the fake samples produced by the generator to learn a sharper decision boundary between different classes from a few samples.  These methods hallucinate samples from only a few examples without modeling the latent class distribution, making the augmented data extremely biased.  As a comparison, our method transfers intra-class generalizable information from many-shot classes to few-shot classes to assist in generating more diverse samples.  cCov-GAN~\cite{gao2018low} leverages a similar idea by preserving covariance information learned from many-shot classes to help better augmentation.  However, the learning of cCov-GAN is independent of downstream tasks, while in our method, the generator is trained end-to-end to ensure the improvement of the downstream classification task.

\section{Method}
\label{sec:method}

\begin{figure*}[t]
\centering
 \includegraphics[width=\textwidth]{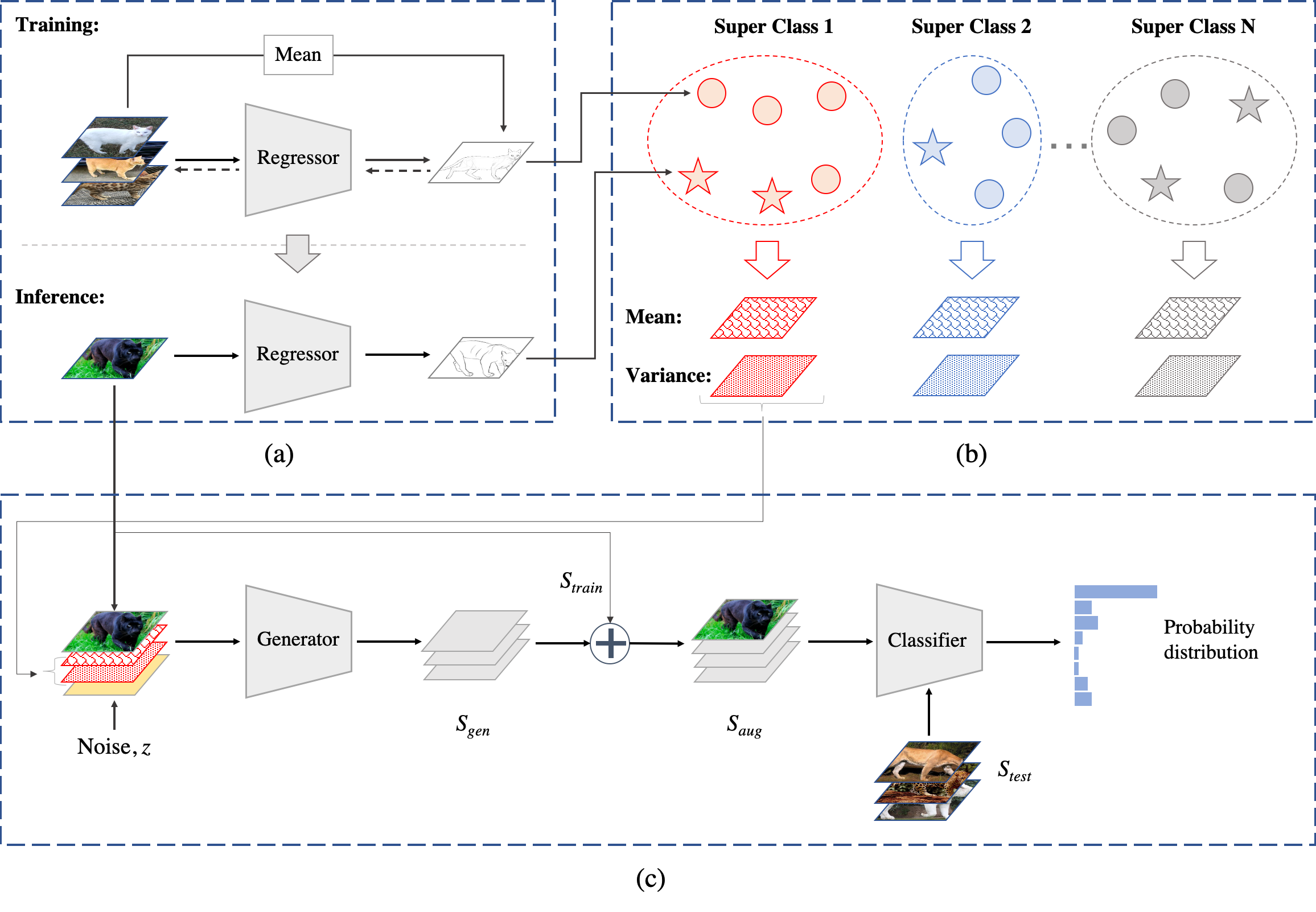}
 \caption{Our approach contains three modules.  (a) presents the class center regression module.  A neural network regressor is trained on many-shot classes and can be used to infer the class means for few-shot classes.  (b) presents the knowledge transferring module.  Superclasses are found using clustering algorithms on the class means, and the few-shot classes will inherit the mean and feature variance of their superclasses. (c) presents the meta-generator module. A generator is used to augment data for few-shot classes. Then, the classifier is trained on the augmented data.  The generator learns under a meta-learning framework.\vspace{-10pt}}
 \label{fig:system}
\end{figure*}

Consider a classification task with unbalanced training dataset $\mathcal{D}_{train}$, which contains both many-shot and few-shot classes, $\mathcal{D}_{train}=\mathcal{D}_{train}^{many} \cup \mathcal{D}_{train}^{few}$.  Our goal is to train a discriminative classifier $h$ on the training data, such that it has high classification accuracy on test data $\mathcal{D}_{test}$, which also contains both many-shot and few-shot classes, $\mathcal{D}_{test}=\mathcal{D}_{test}^{many} \cup \mathcal{D}_{test}^{few}$.  Formally, let $(x, y)\in \mathcal{D}_{test}$ be a sample from test set, and $\hat{\mathbf{p}}(x)$ be the estimated probability distribution over the class labels from $h$:
\begin{equation}
    \hat{\mathbf{p}}(x) = h(x, \mathcal{D}_{train}; \mathbf{w})
\end{equation}
in which, $\mathbf{w}$ is the learnable parameters of $h$.  We want $\hat{\mathbf{p}}(x)$ has the highest value at the correct label $y$.  If sample $(x, y)\in \mathcal{D}_{test}^{many}$, this goal is easy to achieve since the $\mathcal{D}_{test}^{many}$ has sufficient training samples.  However, when $(x, y)\in \mathcal{D}_{test}^{few}$, it is difficult since the size of the training dataset is small. 

Many methods address few-shot learning problems under the meta-learning framework~\cite{thrun1998lifelong}. Concretely, they learn shareable knowledge from different tasks during meta-training and develop a mapping from a given task to a certain classification algorithm.  When a new task is given during meta-test, the best classification algorithm is chosen to obtain good results, provided even only very few samples.  Among these methods, the one proposed by Wang \textit{et al.}~\cite{wang2018low} is the one that most close to our method. Specifically, they use meta-learning to learn a generator whose input seeds are the few-shot samples in the training set and random noise.  The generator outputs more samples labeled as the same class as the input few-shot samples.  Then, the small training dataset can be augmented with these generated samples to train a more discriminative classifier.  We call this framework ``meta-generator".

Our method also leverages the idea of ``meta-generator" for solving the few-shot classification problem.  However, instead of using only a few samples for data augmentation, we propose also to use the transferred knowledge from many-shot classes to generate more diverse samples for augmenting the sparse few-shot training dataset.  Figure \ref{fig:system} presents a system-wise overview of our method, which includes three modules.  The first module is a few-shot class mean regression module, presented in Figure \ref{fig:system}a, which estimates the mean of a few-shot class given only she-shot samples.  Figure \ref{fig:system}b presents the second module, the superclass knowledge transferring module, which builds a two-level hierarchical tree of the whole dataset and computes the statistical mean and variance for each superclass.  The third module is the meta-generator, which takes the few-shot samples, the inherited superclass mean and variance, and a random noise as input, and outputs more samples belonging to the same class.  We will explain the details of each module in the following context.

\subsection{Few-Shot Class Center Regression}
\label{subsec:regressor}

Our idea is to transfer useful knowledge from many-shot classes to few-shot classes.  However, only when the many-shot classes and the few-shot classes are close in feature dimension or belong to the same superclass, such a knowledge transfer is useful.  The class means are required in order to calculate the distance between two classes.  For a many-shot class, we can directly obtain a reasonable estimation of the mean from the abundant training samples.  For a few-shot class, such a way will lead to very biased estimation.  To have a fair estimation of the means for the few-shot classes as well, we trained a neural network regressor to map from few-shot samples to the class mean directly.

We only use the many-shot classes to train the regressor.  For $n$-shot learning, we randomly select $n$ samples as the input, and the output is an estimated class mean.  The training label is obtained by directly taking the mean of all the samples in a many-shot class.  Once the regressor is trained, the mean of a few-shot class can be estimated using the $n$-shot samples as the input.

\subsection{Superclass Knowledge Transferring}

So far, we have the class means computed from the previous module.  We then use the class means to find the underlying connections between classes.  First, we run a $K$ nearest neighbor algorithm over the classes means to cluster the classes into superclasses.  In the experiment, we set $K=5$ by cross-validation.  As a second step, we then compute the class mean and feature variance for each superclass, using all the samples (including both many-shot and few-shot classes) belonging to this superclass.  Finally, we pass the superclass mean and variance from the superclass to its few-shot class children as a shareable knowledge transfer.  The inherited mean and feature variance will be feed into the meta-generator as additional information to augment the data for few-shot classes.

\subsection{Meta-Generator for Data Augmentation}

The last module in our method is the meta-generator.  Specifically, we train a generator under the meta-learning framework, such that the generator can augment additional samples for few-shot classes.  Figure \ref{fig:system}c presents the whole process.  The input of the meta-generator $G$ consists of 3 parts, a random selection $x$ from the few-shot samples, the mean $\mu$, and feature variance $\sigma$ inherited from the superclass, a random noise $z$ which ensure the diversity of the generated samples.  We can thus write the the meta-generator as a parameterized function, $G(x, \mu, \sigma, z; \theta)$, in which $\theta$ is the learnable parameters of $G$.  The output of $G$ is a generated sample $\hat{x}$ with the same dimension and the same label $y$ as the input sample $x$:
\begin{equation}
    \hat{x} = G(x, \mu, \sigma, z; \theta)
\end{equation}
Following, we will explain the meta-training and meta-test processes of the generator.

\subsubsection{Meta-Training}

At each meta-training loop, we first sample a training subset $S_{train}$ and a test subset $S_{test}$ from the whole datatset $\mathcal{D}_{train}$ for both many-shot and few-shot classes.  Let $S_{train}^i \subseteq S_{train}$ be the sampled subset of class $i$, $i \in {1, ..., M}$, in which $M$ is the total number of classes in $S_{train}$.  At each sampling step,  we make sure $|S_{train}^i| \leq n$.  Namely, we want the training subset contains no more than $n$-shot samples for each class.  In the next step, we use the generator to generate an augmentation dataset $S_{gen}$ from $S_{train}$, and then combine the two dataet together to obtain an augmented training dataset $S_{aug} = S_{gen}\cup S_{train}$.  Let $S_{aug}^i \subseteq S_{aug}$ be augmented training dataset of class $i$.  We make sure $|S_{aug}^i| = n_{aug}$ for all $i \in {1, ..., M}$.  Namely, we augment the data for all the classes to reach the same size $n_{aug}$.  Thus, $S_{aug}$ is a balanced dataset.

Once we have the balanced training dataset $S_{aug}$, we first training the classifier $h$ with the loss:
\begin{equation}
l_h = \sum_{(x_i, y_j)\in S_{aug}} L_{CE}(h(x_j, \mathbf{w}), y_j)
\label{eq:loss_h}
\end{equation}
in which, $\mathbf{w}$ is the learnable parameters of $h$, $L_{CE}$ represents the cross-entropy loss, $(x_i, y_j)$ is a sample from $S_{aug}$.  We train the classifier $h$ until convergence inside each meta-training loop.

Once we have a trained classifier $h$, we keep its parameter $\mathbf{w}$.  Then, we use the sampled meta-test dataset $S_{test}$ to measure the performance of the classifier $h$.  The loss we use is in a similar format:
\begin{equation}
    l_g = \sum_{(x_k, y_k)\in S_{test}} L_{CE}(h(x_k, \mathbf{w}), y_k)
    \label{eq:loss_g}
\end{equation}
We compute gradient from the loss in Eq \ref{eq:loss_g} and use it to guide the learning of the generator.

\subsubsection{Meta-Test}

In meta-test, we have a learned generator $G$, a support few-shot dataset $S_{supp}$, and a query dataset $S_{query}$. Let $S_{supp}^t \subseteq S_{supp}$ be the subset of class $t$ in $S_{supp}$.  For each class $t$, we first obtain the class mean $\mu_t$ and variance $\sigma_t$ using the first two modules of our proposed method.  Next, to augment $S_{supp}^t$, we input a randomly selected sample $x_t \in S_{supp}^t$, the class mean $\mu_t$ and variance $\sigma_t$, as well as a random noise $z$ to the trained meta-generator to generate more samples for class $t$.  Let $S_{supp}^{aug}$ be the augmented set of $S_{supp}$.  We then use $S_{supp}^{aug}$ to train the classifier $h$ under the guidance of the loss in Eq \ref{eq:loss_h}.  Once $h$ is trained, we can use it the predict the probability distribution on the label for the query dataset $S_{query}$.  Note that, in meta-training, parameters of the generator are kept fixed.

\section{Experiment}

\subsection{Setup}

\textbf{Datasets }  We perform our experiments starting with the \textit{MiniImageNet} benchmark~\cite{vinyals2016matching}.  MiniImageNet few-shot learning benchmark is a randomly selected subset of ImageNet, consisting of $60,000$ images with 100 classes, each having 600 examples.  The dataset is split into 64 classes for training, 16 classes for validation, and 20 classes for the testing.

We then test our method on a more challenging dataset, \textit{ImageNetFewShot}~\cite{hariharan2017low}, which has 1000 classes with significant intra-class variation. We follow the data splitting strategy in~\cite{hariharan2017low}, dividing the 1000 ImageNet categories randomly into 389 base categories and 611 novel categories.  The base classes are further divided into two subsets $C_{base}^1$ (193 classes) and $C_{base}^2$ (196 classes) and the novel classes into $C_{novel}^1$ (300 classes) and $C_{novel}^2$ (311 classes).  The joint set $C^{cv} = C_{base}^1 \cup C_{novel}^1$ is then used for meta-training and validation, while $C^{fin} = C_{base}^2 \cup C_{novel}^3$ is used for meta-test.

Finally, we conduct another experiment on a long-tailed benchmark, \textit{ImageNet-LT}~\cite{liu2019large}, which is sampled from the original ImageNet-2012~\cite{deng2009imagenet} following the Pareto distribution with the power value $\alpha=6$.  ImageNet-LT has 115.8K images from 1000 categories.  The number of samples in each class ranges from 5 to 1280, such that the training set follows a long-tailed distribution.

\textbf{Evaluation Metrics }  We use the top-1 classification accuracy as the evaluation metric to measure the performance of our method and the baselines.  We report 1-shot and 5-shot accuracies for MiniImageNet and 1-shot, 2-shot, 5-shot, 10-shot, and 20-shot accuracies for ImageNetFewShot in the main paper.  Result on \textit{ImageNet-LT} (1-shot) is reported in the supplementary material.

\textbf{Implementation Details} Our model is implemented using PyTorch\cite{NEURIPS2019_9015}, optimized with stochastic gradient descent method. The initial learning rate is set to be 0.01.  During training, we apply weight decay with the momentum of 0.9 along with a learning rate scheduler, and the multiplicative factor is set to be 0.2.  We train the networks for 60000 iterations in each experiment. All the experiments are done on a single GTX 2080Ti GPU card. For the hierarchy generation, in miniImageNet we used 5 super classes and for ImageNet we used 78 super classes. This number was tweaked on the validation set of the corresponding datasets.

\subsection{Results}

\textbf{Result on MiniImageNet Benchmark}  Table \ref{tab:miniimagenet} presents the results of both our method and the baseline methods on the MiniImageNet Benchmark.  We observe that our method achieves a comparable result as the baselines for $1$-shot setting, but our result is not the SOTA.  However, for $5$-shot setting, our method outperforms the SOTA result by more than $5$ percents, while the largest performance difference between the baselines is less than $3$ percents.  The result of $5$-shot setting validates the effectiveness of our method in solving few-shot classification problems.  Nevertheless, the result of $5$-shot setting indicates there is still a big potential to keep improving the performance.  Here we give our intuition on why our method does not achieve as remarkable performance for $1$-shot setting.  The class mean is estimated from the few samples using the regressor we introduced in Section \ref{subsec:regressor}.  In $1$-shot setting, the mean is estimated from only one sample, while in $5$-shot setting, each of the five samples will lead to an output from the regressor, and the mean is taken as the average of the five outputs.  Since MiniImageNet does not have enough data to train a good regressor, the estimated mean from only 1 sample would be quite biased, while the mean estimated from 5 samples is more close to the real class mean.  As a consequence, the samples generated using the $5$-shot mean will be distributed in a more similar way to the real distribution for that few-shot class, and the trained downstream classifier will have a better decision boundary.  One interesting future work would be how to more accurately estimate class mean from only a single sample under the setting that the training data is sparse.


\begin{table}
\small
\begin{minipage}[t]{0.45\linewidth}
  \begin{tabular}{lcc}
    \toprule
     & \multicolumn{2}{c}{Top-1 accuracy($\%$)} \\
    \cmidrule(r){2-3}
    Method         & n=1     & n=5\\
    \midrule
    TADAM~\tiny{\cite{oreshkin2018tadam}}    & 58.50 \tiny{$\pm0.30$}    & 76.70 \tiny{$\pm0.30$}\\
    TapNet~\tiny{\cite{yoon2019tapnet}}      & 61.65 \tiny{$\pm0.15$}    & 76.36 \tiny{$\pm0.10$}\\
    MetaOpt-SVM~\tiny{\cite{lee2019meta}}    & 62.64 \tiny{$\pm0.61$}    & 78.63 \tiny{$\pm0.46$}\\
    DC~\tiny{\cite{lifchitz2019dense}}       & 61.26 \tiny{$\pm0.20$}    & 79.01 \tiny{$\pm0.13$}\\
    CAN~\tiny{\cite{hou2019cross}}           & \textbf{63.85 \tiny{$\pm0.48$}}    & 79.44 \tiny{$\pm0.34$}\\
    \hline
    Ours           & 61.32 \tiny{$\pm0.61$}    & \textbf{85.02 \tiny{$\pm0.55$}}\\
    \bottomrule
    \end{tabular}
    \vspace{3mm}
    \caption{Results on MiniImageNet.  Our method is comparable with SOTA on $1$-shot setting and has a big improvement on $5$-shot setting comparing to the baselines.}
    \label{tab:miniimagenet}
\end{minipage}
\hspace{0.05\linewidth}
\begin{minipage}[t]{0.45\linewidth}
  \begin{tabular}{lccccc}
    \toprule
     & \multicolumn{5}{c}{Top-1 accuracy($\%$)} \\
    \cmidrule(r){2-6}
    Method     & n=1     & n=2     & n=5     & n=10     & n=20\\
    \midrule
    MN~\tiny{\cite{vinyals2016matching}}         & 43.6    & 54.0    & 66.0    & 72.5     & 76.9\\
    PN~\tiny{\cite{snell2017prototypical}}         & 39.3    & 54.4    & 66.3    & 71.2     & 73.9\\
    Hallu~\tiny{\cite{wang2018low}}    & 45.0    & 55.9    & 67.3    & 73.0     & 76.5\\
    \hline
    Ours       & \textbf{46.5}    & \textbf{56.2}    & \textbf{68.0}    & \textbf{73.6}     & \textbf{76.9}\\
    \bottomrule
    \end{tabular}
    \vspace{3mm}
    \caption{Results on ImageNetFewShot.  Our method consistently outperforms SOTA methods.  For the most difficult $1$-shot setting, our method achieves a large performance boost.}
    \label{tab:imagenetfewshot}
\end{minipage}
\vspace{-5mm}
\end{table}

\begin{figure*}[t]
\centering
 \includegraphics[width=\textwidth]{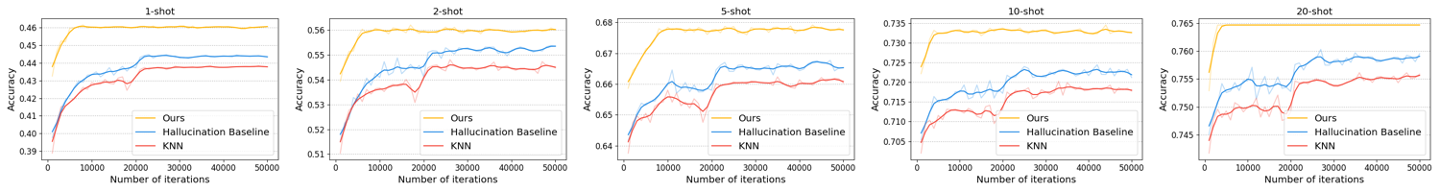}
 \caption{Accuracy plots of our method, the hallucination baseline, and the naive KNN approach on ImageNetFewShot.  Our method consistently outperforms the baseline and the naive KNN approach.  (Zoom in for better visualization.)}
 \vspace{-15pt}
 \label{fig:knn}
\end{figure*}

\textbf{Result on ImageNetFewShot}  Table \ref{tab:imagenetfewshot} presents the results of both our method and the baseline methods on the ImageNetFewShot dataset.  We observe that our method consistently outperforms the baseline methods.  For $1$-shot setting, which is supposed to be the most challenging setting, our method achieved the largest boost comparing to other settings when $n > 1$.  While $n$ gets smaller, the improvement of our method over the Hallucination baseline is larger~\cite{wang2018low}.  Moreover, here our method does not encounter the abnormal performance ``drop'' for the $1$-shot setting.  Different from MiniImageNet, ImageNetFewShot is a much larger dataset with around $14,000$ samples for each many-shot class, while there are only 600 images per class in MiniImageNet.  Because of the sufficient amount of training samples, the regressor trained on ImageNetFewShot can estimate the class mean more accurately, even from only one single sample.  Namely, the class mean estimated from one sample is similar to the class mean estimated from $n$ samples, with $n>1$, and they are both close enough to the real class mean. So, the performance ``drop'' will not happen because of the inaccurate estimation of the class mean.  We would like to point out that our re-implementation Hallucination baseline~\cite{wang2018low} results are around $1$ percent lower for all the settings reported in the paper since we do not know all the implementation details.  In Table \ref{tab:imagenetfewshot}, we still present the results in the original paper.  

\subsection{Ablation Study}

\paragraph{Methods on Finding Intra-Class Information}  As described in Section \ref{subsec:regressor}, we use a deep regressor trained on many-shot classes to map from a single sample to the class mean for few-shot classes.   After that, we cluster the class to superclasses using the means, and use the statistical information of the superclasses as the transferable intra-class information.  However, a more straightforward approach would be directly finding the K-nearest neighbors (KNN) for the few-shot samples, and then using the neighbor classes to help find the superclasses.  We conduct this ablation study and present the result in Figure \ref{fig:knn}.  We set K=5 by cross-validation, and the results in Figure \ref{fig:knn} is on the validation set of ImageNetFewShot~\cite{hariharan2017low}.  We observe that, the naive KNN approach does not even surpass the baseline method, while our method consistently outperforms the baseline.  

\begin{figure*}[t]
\centering
 \includegraphics[width=\textwidth]{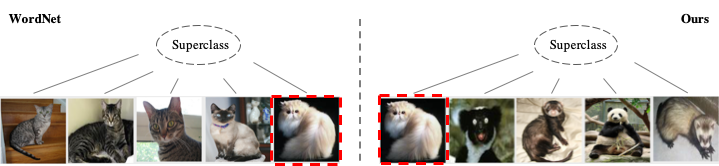}
 \caption{Clustering results using WordNet~\cite{fellbaum2012wordnet} and our method.  WordNet clusters classes that semantically close into the same super category, while our method clusters in the feature dimension.}
 \vspace{-10pt}
 \label{fig:wordnet}
\end{figure*}

\begin{figure}[t]
\centering
    \includegraphics[width=\linewidth]{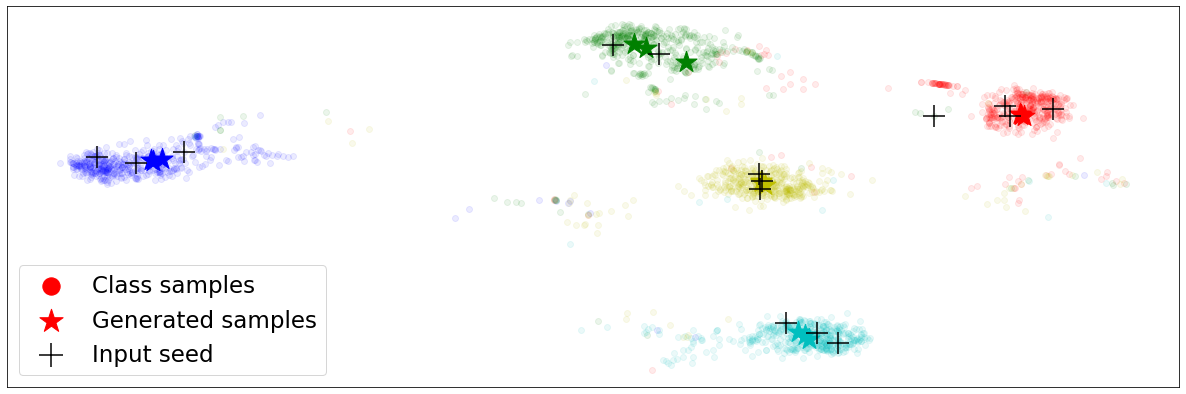}
    \caption{Visualization of class samples, input seed samples for the generator, and generated samples. Generated samples are distributed close to the class mean providing better prototypes to the classifier.}
    \vspace{-15pt}
    \label{fig:class_tsne}
\end{figure}

\vspace{-7pt}
\paragraph{Methods on Building Superclass Hierarchical Tree} Instead of building the superclass tree using the two-step mechanism described in Section \ref{sec:method}, We also tried to use the semantic information to construct the tree.  Specifically, we use WordNet~\cite{fellbaum2012wordnet} to cluster the classes that semantically close to each other to the same superclass.  However, we found that the clustering results from WordNet are extremely imbalanced, leading to bad classification accuracy.  Figure \ref{fig:wordnet} presents an example of the WordNet clustering and our clustering.  We observe that WordNet categorizes persian cat to the same super-category as other kinds of cats. At a comparison, our method performs clustering in the feature dimension and categorizes persian cat and other animals to the same superclass.  When using the clustering result from WordNet, our method performs disastrously.  We believe it is because the semantic feature is different from the feature used by the classifier to perform discrimination.  Thus, our class mean regressor and superclass clustering are both conducted in the feature dimension to keep consistent with the downstream classifier.

\vspace{-7pt}
\paragraph{Inherited Information for Data Augmentation} Figure \ref{fig:class_tsne} shows that the augmented samples belong perfectly to the same classes as the input seeds while being away from the decision boundaries.  The result indicates that the statistical information inherited from super classes can help ensure that the generated samples belong to the same class as the seed. This is because the classes belonging to the same super class share similar statistics and thereby transferable information.  Such statistical information can be treated as the class prototype.  The prototype tells the generator what the object should roughly look like, and the few-shot samples will give the generator detailed information to fine-tune the generated samples.  The result also shows that the regressor trained on many-shot classes can predict meaningful class means for few-shot classes.
\vspace{-7pt}

\section{Conclusion}

In this paper, we proposed a two-step mechanism to extract generalizable intra-class information, which can be transferred from many-shot data to few-shot data.  We leverage such intra-class information transferring to help augment the sparse few-shot data using a generator guided end-to-end by the classification loss.  Our approach achieves state-of-the-art performance on the MiniImageNet benchmark $5$-shot setting and the ImageNetFewShot dataset.  We hope our work can offer some inspiration for future works in solving few-shot learning tasks.


{\small
\bibliographystyle{ieee}
\bibliography{egbib}
}

\newpage
{ \LARGE \textbf{Supplementary Material} }
\setcounter{section}{0}
\section{Algorithm}

As a complementary to the Experiment section, we present here Algorithm \ref{alg:train_alg}, the training algorithm of our proposed method, as a better reference for the readers.  We include details on data pre-processing, mathematical definitions, and some implementation details in Algorithm \ref{alg:train_alg}. 

\begin{algorithm}[H]
\caption{Training Algorithm}
\label{alg:train_alg}
    \begin{algorithmic}[1]
    \Require{\textbf{1)} Training dataset $\mathcal{D}_{train}=\{\mathcal{D}_{train}^{many}, \mathcal{D}_{train}^{few}\}$;  \textbf{2)} Regressor $f(\cdot ; \phi)$;  \textbf{3)} Generator $G(\cdot ; \theta)$;  \textbf{4)} Classifier $h(\cdot ; \mathbf{w})$}
    
    \Ensure{\textbf{1: Class center regression}}
    \State Fit $\phi$ using $\mathcal{D}_{train}^{many}$ to obtain $\phi^*$;
    \State Predict the cluster mean: $\{\mu^i\}$, where $\mu^i=f(\mathcal{D}_{train}^i; \phi^*)$ is the mean of class $i$, $i=1, ..., N$;
    
    \Ensure{\textbf{2: Balanced superclass clustering}}
    \setcounter{ALG@line}{0}
    \State Cluster $\{\mu^i\}$ into $N_{sup}$ superclasses using $k$-NN, compute the class means $\{\mu_{sup}^j\}$, $j=1, ..., N_{sup}$;
    \State For each super class $j$, assign the closest class $k$, $k=1, ..., N$, as its subclass, and repeat the process for $\frac{N}{N_{sup}}$ times;
    \State Compute the variance for each superclass, $\sigma_{sup}^j$
    
    \Ensure{\textbf{3: Meta-generator training}}
    \setcounter{ALG@line}{0}
    \For{Each meta-learning loop}
    \State Sample a low-shot support dataset $\mathcal{S}_{train}$ and a query dataset $\mathcal{S}_{test}$ from the training data $\mathcal{D}_{train}$;
    \State Use the generator to augment the support dataset $\mathcal{S}_{aug}=\mathcal{S}_{train}\cup\{..., (\hat{x}_k, \hat{y}_k), ...\}$, where $\hat{x}_k = G(x_k, \mu_k, \sigma_k, z; \theta)$, $\hat{x}_k$ is the generated sample, $(x_k, y_k)\in\mathcal{S}_{train}$, $\hat{y}_k=y_k$, $(\mu_k, \sigma_k)$ are inherited from the according superclass, $z$ is a noise;
    \State Fit $\mathbf{w}$ using $\mathcal{S}_{aug}$ to obtain $\mathbf{w}^*$;
    \State Fit $\theta$ using $\mathcal{S}_{test}$, guided by the loss, $l_g = \sum_{(x_k, y_k)\in S_{test}} L_{CE}(h(x_k, \mathbf{w}^*), y_k)$, $L_{CE}$ is the cross entropy loss.
    \EndFor
    
    \end{algorithmic}
\end{algorithm}

\section{More Ablation Studies}
We present here some more ablation studies to validate the contribution of each module in our methods.  Figure \ref{fig:suppl_ablation} summarizes the results.  All the experiments are conducted on the ImageNetFewShot dataset~\cite{hariharan2017low} 1-shot setting, and the given plots are on validation dataset.    In each experiment, we change one design that could potentially impact the performance of our method, and compare with our method (final version), as well as the Hallucination baseline~\cite{wang2018low}.

\subsection{Impact of Class Mean Regressor}

\begin{figure*}[t]
\centering
 \includegraphics[width=\textwidth]{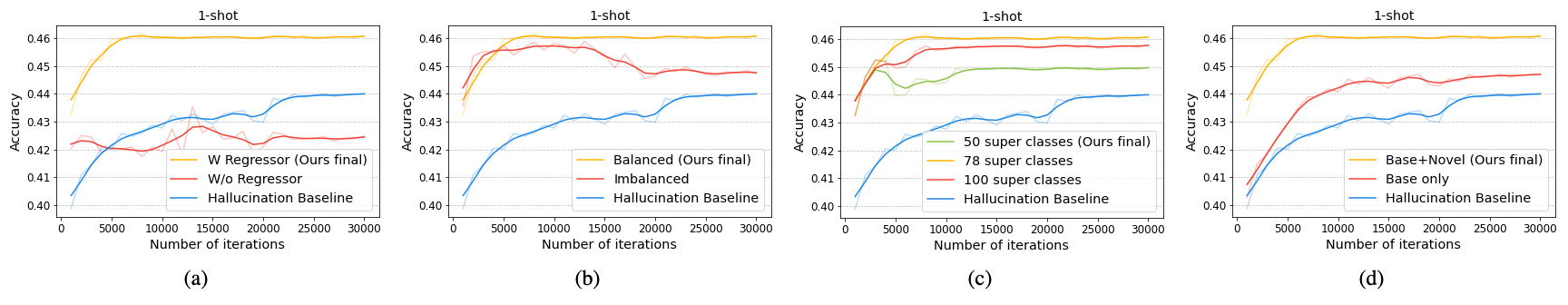}
 \caption{More ablation studies. (a) shows that using the regressor to estimate the class mean achieves much better performance than directly using the 1-shot sample as the class prototype; (b) shows that using the balancing strategy when constructing the superclass tree boosts the performance; (c) presents results with respect to the number of superclasses, and based on these results we set the value to be 78 in all our experiments;  (d) shows that using both base and novel classes to construct the superclass tree achieves a significantly better result than just using the base classes to construct the superclass tree.}
 \label{fig:suppl_ablation}
\end{figure*}

Figure \ref{fig:suppl_ablation}a presents the result of not using the class mean regressor.  Instead, we directly use the 1-shot sample as the estimation of the class mean (or prototype) for clustering.  We observe that the result is much worse than the baseline, as well as our final version, which uses the regressor to predict the class mean for clustering.  This experiment validates the necessity of using the class mean regressor.

\subsection{Superclass Tree Balancing}

The second module of our method is superclass clustering.  However, directly running a $K$-NN over the class means will lead to an imbalanced superclass tree.  To measure the impact of using an imbalanced tree, we conduct this experiment.  Figure \ref{fig:suppl_ablation}b presents the result of using the naive $K$-NN without tree balancing.  We observe that our method, which leverages a tree balancing strategy~\cite{teja2018hydranets}, achieves better performance than the version that directly uses the imbalanced superclass tree.

\subsection{Number of Superclasses}

The number of superclasses is a hyper-parameter which needs to be carefully tuned in order to obtain a better result.  Figure \ref{fig:suppl_ablation}c presents results of using several different values (50, 78, 100) as the superclass number.  We observe that number of superclasses does have an impact on the performance of our method.  In the experiments, we set the number of superclasses to be 78 based on this experiment result.  We did not go further on finding the global optimal value, but we do believe the performance could be further improved with a better superclass number design.

\subsection{Superclass Tree Construction with Only Base Classes}

Figure \ref{fig:suppl_ablation}d presents the result of constructing the superclass tree using only the base (or many-shot) classes. Explicitly, we compute the mean for each base class, cluster the class means into superclasses, and calculate the superclasses means.  As a second step, we assign each novel (or few-shot) class to the closest superclass.  We observe from the result that the version using only the base classes for superclass tree construction performs worse than our final version and the baseline.  This experiment reveals that using the few-shot samples to estimate the class mean (or a prototype) is necessary.

\section{Evaluation on Long-Tailed Dataset}

We also test the generalizability of our method with a long-tailed dataset, \textit{ImageNet-LT}~\cite{liu2019large}, which is sampled from the original \textit{ImageNet-2012}~\cite{deng2009imagenet} following the Pareto distribution with the power value $\alpha=6$.  \textit{ImageNet-LT} has 115.8K images from 1000 categories.  The number of samples in each class ranges from 5 to 1280, such that the training set follows a long-tailed distribution.

\begin{table}[h]
  \caption{Results on ImageNet-LT benchmark.}
  \label{tab:imagenetlt}
  \centering
  \begin{tabular}{lccc}
    \toprule
     & \multicolumn{3}{c}{Top-1 accuracy ($\%$)} \\
    \cmidrule(r){2-4}
    Method                                               & Many   & Medium          & Few\\
    \midrule
    Prototypical Networks~\cite{snell2017prototypical}                   & 61.35  & 35.16           & 12.86 \\
    Ours                                                 & 61.35  & \textbf{37.57}  & \textbf{18.25} \\
    \bottomrule
  \end{tabular}
\end{table}

We follow exactly the learning procedure described in the main paper and Algorithm \ref{alg:train_alg}.  The baseline is the Prototypical Networks~\cite{snell2017prototypical}, which is directly trained on the long-tailed data without any data augmentation. At the same time, we augment the data for few-shot classes in our method.  We observe that our method significantly boosts the classification accuracy for few-shot classes.  This experiment shows that our method can also be used in improving the performance for long-tailed tasks.

\end{document}